\documentclass[conference]{IEEEtran}

\usepackage{graphicx}
\usepackage{amssymb}
\usepackage{amsmath}
\usepackage{amsfonts}
\usepackage{textcomp}
\usepackage{makeidx}
\usepackage{multirow}
\usepackage{verbatim}
\usepackage[usenames]{color}
\usepackage{algorithm}
\usepackage{algorithmic}

\newcommand{\CASE}[1]{\STATE \textbf{case} #1\textbf{:} \begin{ALC@g}}
\newcommand{\ENDCASE}{\end{ALC@g}}

\newcommand{\DEFAULT}{\STATE \textbf{default:} \begin{ALC@g}}
\newcommand{\ENDDEFAULT}{\end{ALC@g}}
\newcommand{\DEFAULTLINE}[1]{\STATE \textbf{default:} }
\usepackage{array}
\usepackage{eqparbox}

\ifCLASSINFOpdf
\else
\fi
\begin{document}

 \begin{titlepage}
 \begin{center}
 {\Large \sc PREPRINT VERSION\\}
  \vspace{5mm}
{\huge A First Attempt to Cloud-Based User Verification in Distributed System\\}
 \vspace{10mm}
 {\Large M. Wo\'{z}niak, D. Po{\l}ap,  G. Borowik and C. Napoli}\\~\\
 {\large Email: napoli@dmi.unict.it\\}~\\
 \vspace{5mm}
{\Large \sc FINAL VERSION PUBLISHED ON:\\~\\ \bf Asia-Pacific Conference on Computer Aided System Engineering (APCASE), pp. 226--231 (2015)}
 \end{center}
 \vspace{5mm}
 {\Large \sc BIBITEX: \\}
 
@InProceedings\{Wozniak2015afirst,\\
  author =    \{Wozniak, M. and Polap, D. and Borowik, G. and Napoli, C.\},\\
  title =     \{A First Attempt to Cloud-Based User Verification in Distributed System\},\\
  booktitle = \{Asia-Pacific Conference on Computer Aided System Engineering (APCASE)\},\\
  year =      \{2015\},\\
  pages =     \{226--231\},\\
  address =   \{14-16 July, Quito, Ecuador\},\\
  publisher = \{IEEE\},\\
  doi =       \{10.1109/APCASE.2015.47\},\\
  url =       \{http://ieeexplore.ieee.org/xpls/abs\_all.jsp?arnumber=7287024\}\\
\}\\

 \vspace{5mm}
 \begin{center}
Published version copyright \copyright~2015 IEEE \\
\vspace{5mm}
UPLOADED UNDER SELF-ARCHIVING POLICIES\\
NO COPYRIGHT INFRINGEMENT INTENDED \\
 \end{center}
\end{titlepage}

%
\title{A First Attempt to Cloud-Based User Verification \\in Distributed System}

\author{
\IEEEauthorblockN{Marcin Wo{\'z}niak\IEEEauthorrefmark{1}, Dawid Po{\l}ap\IEEEauthorrefmark{1}, Grzegorz Borowik\IEEEauthorrefmark{3}, Christian Napoli\IEEEauthorrefmark{2}}

\IEEEauthorblockA{
\IEEEauthorrefmark{1}Institute of Mathematics, Silesian University of Technology, Kaszubska 23, 44-100 Gliwice, Poland\\ Email: Marcin.Wozniak@polsl.pl, Dawid.Polap@gmail.com}

\IEEEauthorblockA{\IEEEauthorrefmark{3}Warsaw University of Technology, Nowowiejska 15/19, 00-665 Warsaw, Poland \\
Email: G.Borowik@tele.pw.edu.pl}

\IEEEauthorblockA{\IEEEauthorrefmark{2}Department of Mathematics and Informatics, University of Catania, Viale A. Doria 6, 95125 Catania, Italy\\
Email: Napoli@dmi.unict.it}
}


%


\maketitle

\begin{abstract}
In this paper, the idea of client verification in distributed systems is presented. The proposed solution presents a sample system where client verification through cloud resources using input signature is discussed. For different signatures the proposed method has been examined. Research results are presented and discussed to show potential advantages.
\end{abstract}



%
\IEEEpeerreviewmaketitle

\section{Introduction}\label{sec:intro}
Nowadays, signature recognition is one of the most important operations in various agencies. At each step we are required to verify our identity. A few of the most important examples in which we use the signature are financial institutions such as banks or even stores where people use cards or checks that require confirmation of identity. Another example is approving of labor contracts or any other type of transactions. These are real-life examples in which we use signatures. All these systems use distributed computing. When a client sign documents the computer system is using data stored remotely. Therefore intelligent solutions dedicated for this process are required. Confirmation of identity by signature brings the possibility of forgery, which is driving force for many people to create system not only to confirm but also to detect a possible attempt to forge a signature. It is not an easy task because a person can sign at a certain angle, invert the lines, write letters quickly or even with some defects such as trembling hand. 

Distributed computing system to signature verification should be able to counter check in each of mentioned situations. This is the main problem of decision support systems, which in some way have to process the input signatures to the format or size that allows correct verification. Moreover, mentioned systems should find some specific features in signatures, which are divided into groups describing local and global features. As a global features, the signature is considered as a whole, so for instance a feature will be time or speed of folding. On the other hand, the local will contain characteristic features due to a person's handwriting. In many cases, it is a difficult problem to ask a person to provide many signatures, so it is important to create more signatures in different configuration by computer application on the assumption of saving the most important characteristic features. In recent years, many new ideas for preprocessing and signature verification were presented \cite{cpalka2014_1}, \cite{cpalka2014_2}, \cite{Wozniak2014_2}.
\subsection{Related Works}
Examinations of distributed systems concern workflow management \cite{Wozniak2015_1}. This is very important for overall performance of processing data over cloud resources \cite{napoli2014cloud}, \cite{capizzi2012aninnovative}, \cite{GiuntaPT11}, \cite{Wozniak2014_9}. Another aspect is proper positioning for request management \cite{Wozniak2015_4}, \cite{Wozniak2014_8}, \cite{Wozniak2014_1}, \cite{Wozniak2014_10}, \cite{Wozniak2013_3}, which makes the system faster and more reliable. In various aspects of positioning and optimization Computational Intelligence (CI) provides interesting solutions. CI methods are applicable in various complex problems \cite{Wozniak2015_1}, \cite{Signal-2013}, \cite{Borowik-2009}, \cite{Borowik-2011a}, dynamic systems positioning \cite{Wozniak2014_13} and image processing \cite{Wozniak2014_5}, \cite{Wozniak2014_6}, \cite{Wozniak2015_3}. Therefore proposed solution of signature verification in distributed system will use CI method for intelligent data aggregation. The importance of the presented solution is to efficiently assist in signature verification in distributed systems. The novel approach is based on Cloud-Computed knowledge for various users machines that are about to verify input signatures form serviced clients.
\section{General Idea of Cloud-Computing}\label{sec:cloudComputing}
Cloud-Computing is becoming more popular in the recent years. It is a widely offered service, which basis is a shift of all IT services to the external provider (server) what creates a permanent, remote access to the services for various users located in remote positions. It may use an agent, which is an encapsulated computer system located in the special environment to serve as a core station for data processing, which role is an automatic operations management \cite{Wozniak2015_2}. 
 
Cloud-Computing is often used by companies in order to provide an efficient service to the clients and the users (workers). Instead of creating an infrastructure in all company branches and hiring employers to manage it, the headquarters use central infrastructure and computing power to offer dedicated service for users and clients. In \cite{sanaei2014heterogeneity} authors analyzed Mobile Cloud Computing in terms of heterogeneity. In Cloud-Computing an important issue is security \cite{vollmar2014hypervisor}, \cite{zissis2012addressing}, because users should not be allowed to integrate activities and their resources should be protected from other, unwanted persons.

Computing power can increase and decrease according to the customer needs in specific time. However tailored system composition and application of dedicated methods can improve the overall efficiency and reduce energy consumption \cite{bonanno2014optimal}, \cite{bonanno2012optimal}, \cite{bonanno2011hybrid}. According to research conducted for the European Commission in $2011$, Cloud-Computing can decrease amount of money spend on IT by $20\%$ and lower energy consumption. Cloud-Computing is usually used in one of three ways:
\begin{itemize}
	\item Infrastructure as a Service (IaaS), model in which a special infrastructure is available to the users.
	\item Platform as a Service (PaaS), model in which a special platform of virtual machines and operation systems is available for users to work in.
	\item Software as a Service (SaaS), where users get an access to applications without integrating into system.
\end{itemize}
However mainly some compositions of the mentioned above are most practical. In the article we discuss a combination of IaaS and SaaS applied together to serve as an infrastructure for distributed user requests processing and intelligent software for verification of the processed requests.
\subsection{Applied Model of Processing}\label{sec:distributed}
\begin{figure*}[!th]
\centering
\includegraphics[width=0.7\textwidth]{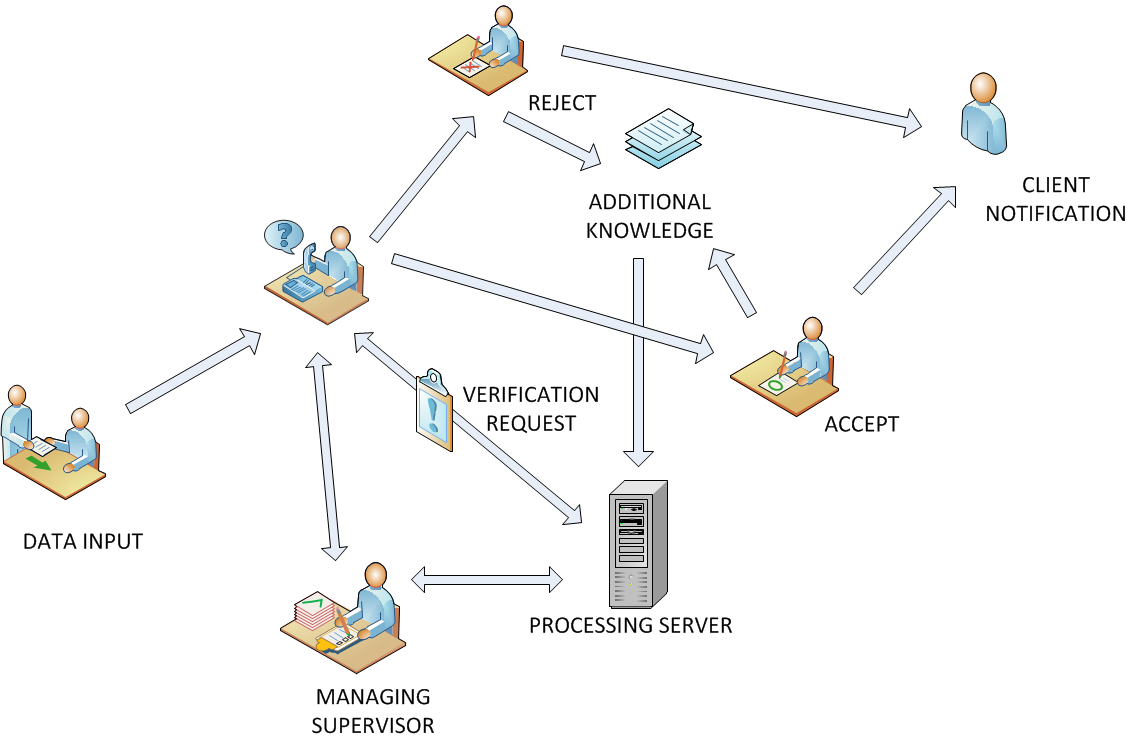}
\caption{Proposed model of requests processing for client verification in Cloud-Computing system.}
\label{fig:processingmodel}
\end{figure*}

\begin{figure*}[!th]
\centering
\includegraphics[width=0.7\textwidth]{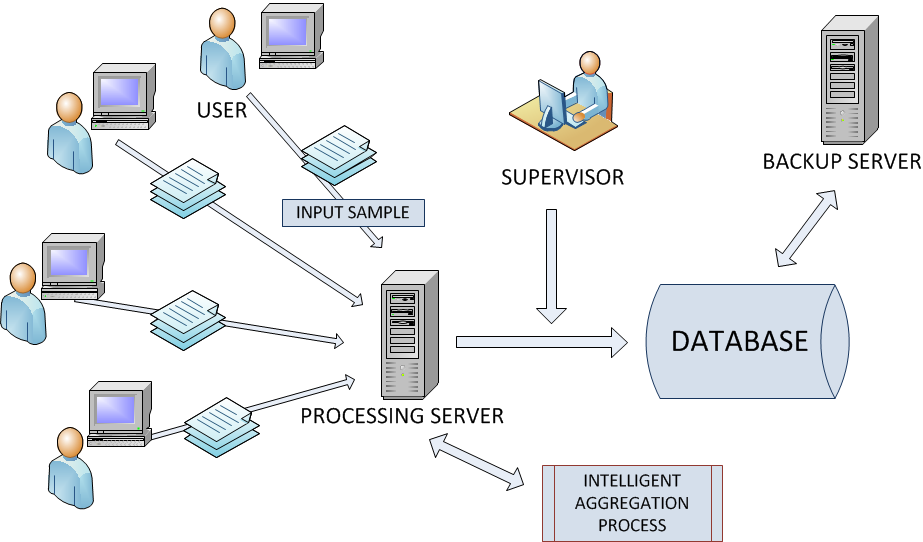}
\caption{Proposed model of knowledge acquisition for future client verification in Cloud-Computing system.}
\label{fig:distributedmodel}
\end{figure*}

In the proposed solution cloud architecture is composed. There are defined two models: one for verification request processing (as shown in Fig. \ref{fig:processingmodel}) and the other for knowledge acquisition process (as shown in Fig. \ref{fig:distributedmodel}).

Request processing (Fig. \ref{fig:processingmodel}) in the system starts when a user of the system wants to verify personality of the client signing documents. The client machine sends the verification request to the processing server. This computing unit is connected to database of signature models. The request is processed (verified) using this knowledge, however recognition goes along with assistance of managing supervisor. This does not mean that all the verified requests must be accepted by the managing unit, but that if the processing server hesitates, i.e. the acceptance ration is under certain limit, the managing supervisor takes action to judge the request using the knowledge stored in the database. After acceptance or rejection of the request, this knowledge is reused again to improve the database for further processing.

The other part of the system is a model of knowledge acquisition for future clients verification (Fig. \ref{fig:distributedmodel}). The system must have the knowledge proper for verification purposes. Therefore the first operation is to help the system learn about the features to verify. Clients give a set of signatures that are send to processing server as an input sample. This set is processed in intelligent aggregation to establish knowledge stored in database. This knowledge can be improved any time, after each successful client verification the new signature is sent to the processing server as a new input sample to be stored in the database after intelligent aggregation with previous knowledge. That means, each new client verification is an important possibility to extend the knowledge in the distributed system. Therefore the system not only performs verification but also learns all the time. Learning process is supervised by an experienced workers to ensure that only the important information helpful for further verification is stored in the database. The system is equipped in backup server, which is to store backup of the knowledge and control the process in the main server. This type of backup ensures that any operation in the system is easily erased and the knowledge can be restored if any demand can happen. For the verification purposes main role plays intelligent signature aggregation method. It is responsible for analyzing input signatures and composing an aggregated knowledge.
\section{Intelligent Aggregation Method}\label{sec:aggregation}
Generally, the signature is a figurative mark containing the name of the signatory. In the electronic version, it is an image made up of black and white pixels where black ones represent the signature. Each pixel can be treated as a given discrete two-dimensional point, and that is why we can use an idea of creating new samples based on set of several samples aggregated into one shape. Proposed system process a few signatures at the same time transforming them into a single form using the method described in \cite{Wozniak2014_2} and place this aggregation into the database. Aggregated structure is perfect for knowledge transfer over connection in distributed systems due to low size of the knowledge portion. The method is presented in Algorithm \ref{alg:Model}. Processing model \cite{Wozniak2014_2} is based on interpolation. At the beginning signature is simplified - for all vertical points one as an arithmetic sum of all them is calculated. After that operation, all created points from simplified signature are subjected to the chosen interpolation method. In the end, graph of the interpolated function is created as a new simplified sample.

\begin{figure*}[!th]
\centering
\includegraphics[width=0.65\textwidth]{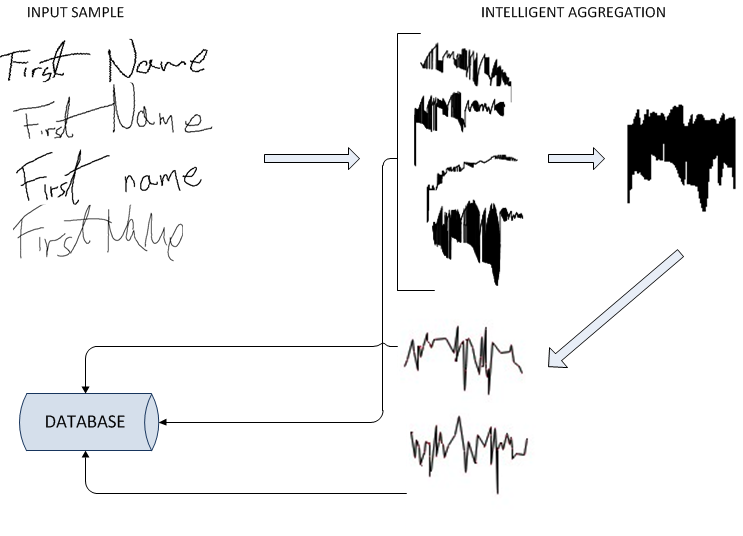}
\caption{Proposed model of intelligent input data aggregation for the knowledge applied in Cloud-Computing based verification process.}
\label{fig:model}
\end{figure*}

Next step is a new approach to create more samples based on existing ones. In the situation, when client is creating a few samples of his/her signatures for the first time, it would be stored in database, and then signatures are to be duplicated in various combinations to improve the knowledge in the database. At first, all samples would be combined to create one sample with a complete area of black pixel. In this area, basis points would be created using Simulated Annealing described in section \ref{sec:annealing}. Then, according to the position of basic points on the $x$-axis, these points should be connected. As a result of that operation, new aggregated sample is created and added to the database for further verification. The model is presented in in Fig. \ref{fig:model} and discussed in Algorithm \ref{alg:Model}. 
\begin{algorithm}[!ht]
\caption{Proposed Signature Processing Model}
\label{alg:Model}
\begin{algorithmic}[1]
\STATE Get several signatures from the client and send it to the processing server in the input sample,
\STATE $k$ is the number of signatures in the input sample,
\STATE	$t:=0$,
	\WHILE{$t \leq k$}
		\STATE Simplify the input sample,
		\STATE Define a limited area,
		\STATE Find basis points using Simulated Annealing presented in Algorithm \ref{alg:SA},			
		\STATE Create a new aggregated sample based on these points,
		\STATE Add new aggregation to the database for further verification,
		\STATE Next iteration $t++$,
	\ENDWHILE
\end{algorithmic}
\end{algorithm}
\subsection{Simulated Annealing}\label{sec:annealing}
Annealing is a metallurgical process where first metal heating to a high temperature is performed, then instantaneous annealing which means holding the metal at a given temperature and in the end slow cooling. The cooling rate is essential in this process. Supervising cooling keeps the metal state close to thermodynamic equilibrium, i.e. the state in which parameters such as pressure and volume remain constant in time. However to maintain a balance we must be aware of the following components:
\begin{enumerate}
	\item thermal equilibrium - the constant temperature with no heat exchange with the environment,
	\item mechanical equilibrium - pressure at any point is constant,
	\item chemical equilibrium - there are no chemical reactions which could change the composition of the metal.
\end{enumerate}
Basic equation which is used to describe the thermodynamics of the process is 
\begin{equation}\label{eq:thermodynamics1}
P(E)\approx e^{-\frac{E}{kT}},
\end{equation}
where $E$ is a thermodynamic system, $k$ is Boltzmann's constant, and $T$ means absolute temperature.

In the metallurgical industry there are several types of that process, for instance homogenizing or plasticizes. The first one involves heating the metal to a temperature of $1000^\circ C-1200^\circ C$, then cooling in the air. The advantage of this is reduction of the crystallization effect. This process has been modeled for optimization purposes.

\subsubsection{Mathematical Model}\label{sec:annealingmodel}
Simulated Annealing (SA) algorithm for the first time was presented in $1983$ by Kirkpatrick and others \cite{Kirkpatrick1983}. In \cite{Ingber1989} authors presented the possibility of modifying the algorithm by introducing a further annealing which will speed up the algorithm. The algorithm was also used in other areas, authors of \cite{Dupanloup2002} showed possibility of using SA to define populations models. In \cite{Svergun1999} the use of SA in quantum chemistry was demonstrated.

The algorithm assumes that the temperature at the beginning of the process is high, which allows frequent changes in configurations. Choosing a solution depends on a certain probability and that is why there exist possibility of choosing the worst solution. The advantage of this solution is the situation when algorithm is located in local optimum - the choice of worse solution makes it possible to leave the optimum and further searching of the global optimum. If the temperature is lower, the probability of choosing a worse solution is smaller, so it is a criterion for the acceptation of the solution. SA uses a modified \eqref{eq:thermodynamics1} in the form
\begin{equation}\label{eq:termodynamika2}
P(E)\approx e^{-\frac{\delta}{T}},
\end{equation}
where $\delta$ is the difference between the value of the fitness function of the current solution $\textbf{x}$ and the new, random solution $\textbf{x}'$ from neighborhood of $\textbf{x}$, which is presented in 
\begin{equation}\label{eq:roznica}
\delta=f(\textbf{x}')-f(\textbf{x}).
\end{equation}
The criterion of acceptance of the new solution is
\begin{equation}\label{eq:krytAkcep}
\gamma < e^{-\frac{\delta}{T}},
\end{equation}
where $\gamma \in (0,1)$ is random value. Other important aspect of the algorithm is a temperature function and stop criterion. The change in temperature might be defined similarly to authors of \cite{Corana1987}, \cite{Svergun1999} by introducing the following equation
\begin{equation}\label{eq:temp}
T_{k+1}=r*T_k,
\end{equation}
where $T_{k}$ means value of the temperature in $k$-th iteration, and $r$ is constant defined before the beginning of the algorithm.

The presented algorithm is an iterative method so it is important to define a stop criterion - the number of iterations or minimum temperature. These are important parameters, as many iterations will lower probability to find best solution. Simulated Annealing is presented in Algorithm \ref{alg:SA}.
\begin{algorithm}[!ht]
\caption{Simulated Annealing}
\label{alg:SA}
\begin{algorithmic}[1]
\STATE Define the value of the initial temperature $T$, the fitness function $f$, the number of iterations $it$ and the temperature change $r$,
\STATE Generate a random initial solution $\textbf{x}$,
\STATE k:=0,			
	\WHILE{$k \leq it$}
		\STATE i:=0,
		\FOR{i \TO k}
			\STATE Generate random neighboring solution $\textbf{x}'$,
			\STATE Calculate the difference $\delta$ using \eqref{eq:roznica},
			\IF{$\delta<0$}
				\STATE $\textbf{x}=\textbf{x}'$
			\ELSE
				\STATE Generate a random value $\gamma$,
				\IF{equation \eqref{eq:krytAkcep} is compiled}
					\STATE $\textbf{x}=\textbf{x}'$,
				\ENDIF
			\ENDIF
		\ENDFOR
		\STATE Reduce the temperature using \eqref{eq:temp},
		\STATE Increase the iterator variable $k:=k+1$,
	\ENDWHILE
\STATE Return $\textbf{x}$ from the last iteration.
\end{algorithmic}
\end{algorithm}
\section{Discussion}\label{sec:Discussion}
Many classifiers are designed to verify signatures, and all of them need a huge database, what is not efficient in Cloud-Computing where the amount of data to be transfered over the network can significantly influence workflow. Presented approach to creating new samples based on existing ones can be a strong advantage for distributed systems. Adding to this process a new technology for computers on which verification software is installed will accelerate operations not burdened by typical workflow restrictions.

\begin{figure}[!th]
\centering
\includegraphics[width=0.2\textwidth]{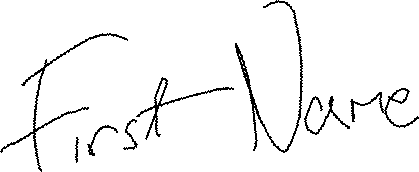}
\includegraphics[width=0.2\textwidth]{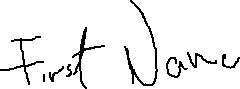}
\includegraphics[width=0.2\textwidth]{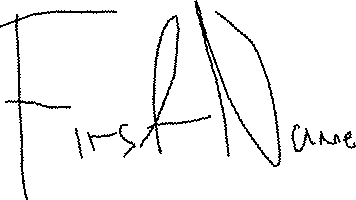}
\includegraphics[width=0.2\textwidth]{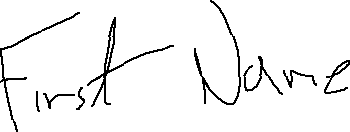}
\caption{Input sample from the client.}
\label{fig:signatureoriginal}
\end{figure}

The proposed intelligent method can manage input signature to make it aggregated for recognition over Cloud-Computing, where by the presented method we can only get the aggregated shape of the clients signature. This aggregated shape is then passed to the database, where it serves for on-line verification and further training. In Fig. \ref{fig:signatureoriginal} a sample set of input signatures is presented, for which proposed intelligent aggregation method evaluated the shape to be passed to the database.
\begin{figure}[!th]
\centering
\includegraphics[width=0.2\textwidth]{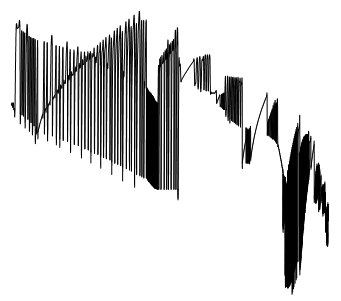}
\includegraphics[width=0.2\textwidth]{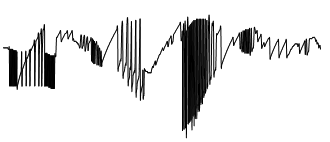}
\includegraphics[width=0.2\textwidth]{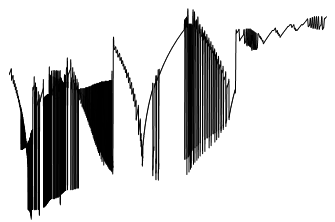}
\includegraphics[width=0.2\textwidth]{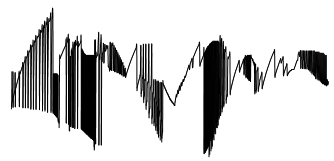}
\caption{Input sample processed in the intelligent system.}
\label{fig:signatureaggregated}
\end{figure}
\begin{figure}[!th]
\centering
\includegraphics[width=0.3\textwidth]{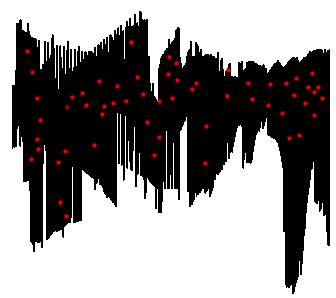}
\includegraphics[width=0.3\textwidth]{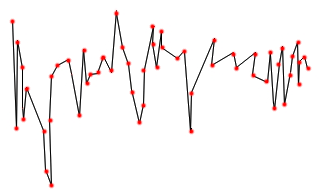}
\caption{Aggregated data with verified important points and the main line for ad-hoc signature verification.}
\label{fig:signaturedecision}
\end{figure}
The input sample is processed using one of the methods presented in \cite{Wozniak2014_2} what gives a set for simplified signatures shapes presented in Fig. \ref{fig:signatureaggregated}. These shapes are aggregated in one final shape that is to present the possible way the client sign documents. This aggregated shape is used for verification, where the SA method shows points crucial for verification. This knowledge goes to database and is used whenever a client has signed a document and system user requests verification. In verification the signature is simplified using methods examined in \cite{Wozniak2014_2} to obtain simplified shape which is compared with important points showed by SA over the aggregated shape. 
\section{Final Remarks}
\label{sec:Conclusion}
In the article we present a novel approach to creating aggregated knowledge for distributed systems. Proposed novel recognition method is easy to implement with possibility to improve. Moreover it can be implemented in Cloud-Computing where low data packages influence workflow stability and performance. Because of that it will make it easier and cheaper for many small companies to use. Since the data package for client verification is aggregated transfer over the network will be fast even for slow connections. 

Future research will lead to define improvements of discussed approach leading to create fast and efficient signature verification system for large distributed systems. 

\section*{Acknowledgements}
This work has been partially supported by project PRISMA PON04a2 A/F funded by the Italian Ministry of University and Research within PON 2007-2013 framework.

\IEEEtriggeratref{99}
\bibliographystyle{IEEEtran}

\bibliography{wozniak_publications_daty,annealing,apcase3}

\end{document}